\documentclass[letterpaper, 10 pt, conference]{ieeeconf}  

\IEEEoverridecommandlockouts                              

\usepackage{hyperref}       
\usepackage{url}            
\usepackage{graphicx} 
\usepackage{amsmath} 
\usepackage{amssymb}  
\usepackage{paralist}

\title{\LARGE \bf
Using Visual Anomaly Detection for Task Execution Monitoring
}

\author{Santosh Thoduka$^{1}$ and Juergen Gall$^{2}$ and Paul G. Pl\"{o}ger$^{1}$
\thanks{This work has been supported by the Bonn-Aachen International Center for Information Technology, a PhD scholarship from the Graduate Institute at Hochschule Bonn-Rhein-Sieg, and the Deutsche Forschungsgemeinschaft (DFG, German Research Foundation) - GA 1927/6-2 (FOR 2535).}
\thanks{$^{1}$Santosh Thoduka and Paul G. Pl\"{o}ger are with the Computer Science department,
         Hochschule Bonn-Rhein-Sieg, Sankt Augustin 53757, Germany
        {\tt\small {santosh.thoduka, paul.ploeger}@h-brs.de}}%
\thanks{$^{2}$Juergen Gall is with the Computer Vision department, University of Bonn,
        Bonn 53111, Germany
        {\tt\small gall@iai.uni-bonn.de}}%
}

\begin{document}
\thispagestyle{empty}
\begin{minipage}[t]{\textwidth}
\textcopyright 2021 IEEE.  Personal use of this material is permitted.  Permission from IEEE must be obtained for all other uses, in any current or future media, including reprinting/republishing this material for advertising or promotional purposes, creating new collective works, for resale or redistribution to servers or lists, or reuse of any copyrighted component of this work in other works.\\
The published version can be found at \url{https://doi.org/10.1109/IROS51168.2021.9636133}.
\end{minipage}
\newpage

\maketitle
\thispagestyle{empty}
\pagestyle{empty}

\begin{abstract}

Execution monitoring is essential for robots to detect and respond to failures.
Since it is impossible to enumerate all failures for a given task, we learn from successful executions of the task to detect visual anomalies during runtime.
Our method learns to predict the motions that occur during the nominal execution of a task, including camera and robot body motion.
A probabilistic U-Net architecture is used to learn to predict optical flow, and the robot's kinematics and 3D model are used to model camera and body motion.
The errors between the observed and predicted motion are used to calculate an anomaly score.
We evaluate our method on a dataset of a robot placing a book on a shelf, which includes anomalies such as falling books, camera occlusions, and robot disturbances.
We find that modeling camera and body motion, in addition to the learning-based optical flow prediction, results in an improvement of the area under the receiver operating characteristic curve from 0.752 to 0.804, and the area under the precision-recall curve from 0.467 to 0.549.
\end{abstract}

\section{INTRODUCTION}
Service robots operating in dynamic, unstructured environments are prone to failures.
In order to be resilient to failures, execution monitoring is necessary, and usually forms a part of the robot's software architecture~\cite{pettersson2005execution}.
Monitoring allows the robot to verify that actions are completed successfully, and perform recovery actions if a failure occurs.
This, in turn, makes robots safer, more trustworthy, and dependable.

A task can have different modes of failure, some of which might be unforeseeable and therefore impossible to enumerate.
However, variations in nominal executions are limited, thereby making them easier to enumerate or model.
Therefore, some authors have framed execution monitoring as an anomaly detection task, in which models are learned from successful executions, and deviations from the models are detected as anomalies~\cite{park2019multimodal, khalastchi2015online}.
Depending on the nature of the failure, different sensors might be used for detection; for example, force-torque sensors for collisions, or vision and auditory sensors for external events such as objects falling.
Some methods~\cite{wellhausen2020safe, mauro2018deep, inceoglu2018failure} only make use of visual data such as RGB frames, depth frames, or the output of object detection algorithms.
In this paper, we propose a method for visual execution monitoring, using the videos from the robot's camera and the robot's kinematics.

\begin{figure}[tpb]
   \centering
   \includegraphics[width=\linewidth]{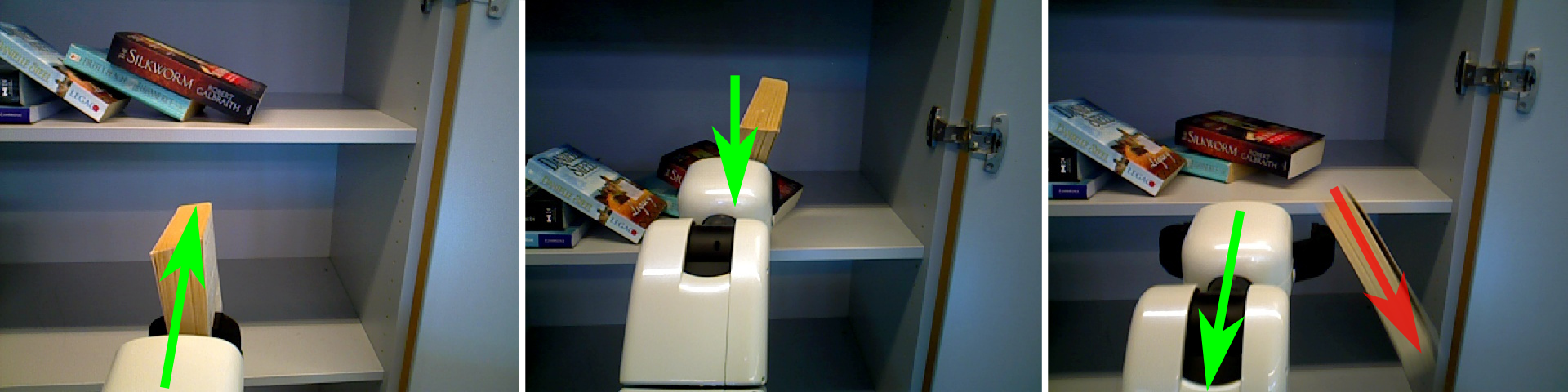}
    \caption{Task failures, such as a book falling while being placed on a shelf, can occur during execution. Modelling expected motions (green arrows) from successful executions allows us to identify unexpected motions (red arrow) using video anomaly detection.}
   \label{fig:introanomaly}
\end{figure}

Research on video-based anomaly detection has focused primarily on surveillance videos, and is driven by datasets such as CUHK Avenue~\cite{lu2013abnormal}, Street Scene~\cite{ramachandra2020street}, etc.
The typical setup for these datasets involves a static camera observing a fixed scene.
For a mobile robot, we can assume neither a static camera nor a fixed scene, since the robot can perform a given task at different locations and the task might require camera motion during execution.
Additionally, the manipulator might occlude parts of the scene and cause motions in the scene which the robot must take into account.

Since deep learning dominates most recent approaches for video anomaly detection, large-scale datasets are needed and have been published in recent years~\cite{ramachandra2020street, sultani2018real}.
Datasets for robotics also exist~\cite{wang2019multimodal,inceoglu2020fino}, and are usually recorded with a single robot, and are task-specific.
Generalizing learning models across robots requires large-scale datasets, which can be quite expensive~\cite{dasari2019robonet}.
Using small, robot- and task-specific datasets sidesteps this problem, but constrains the learning method to be data-efficient.
A combination of model-based and data-driven techniques loosens the requirement for large-scale datasets, while still making it possible to learn aspects which cannot be easily modelled.

For certain actions, robots use predefined manipulator motions or motion primitives, which are only parameterized by a target pose.
This means that the motions resulting from the robot's actions are predictable for a nominal execution.
For the action of placing a book on a shelf, shown in Fig.~\ref{fig:introanomaly}, the dominant motion patterns in a nominal execution include the robot's arm approaching the shelf, a short downward motion when the book is released, and the motion of the arm retracting.
The motion of the book falling off the shelf is not one that occurs during a nominal execution, and therefore should be detected as an anomaly.

In this paper, we learn a model of nominal motions during the execution of an action, and detect anomalies by comparing the observed motion to the expected nominal motion.
The motion of the camera and the robot's body are modelled using the known joint states and kinematics, while motions external to the robot are modelled by learning to predict future optical flow in a self-supervised manner.
The combination of a learning-based video anomaly detection method and analytical modelling of the robot's self motion is in contrast to existing methods which either use only learning~\cite{park2019multimodal}, or use model knowledge at a later stage (for e.g. to isolate faults~\cite{khalastchi2018sensor}).
For evaluation, we collect a dataset of the Toyota HSR robot placing a book on a shelf, which consists of RGB images from the robot's head camera, and joint states.
Both nominal and failed executions are recorded, but the training data consists only of nominal data.\footnote{Code and dataset are available at \url{https://sthoduka.github.io/motion_anomaly_detection/}}
%

\section{RELATED WORK}
\subsection{Execution monitoring in robotics}
Our approach can be considered a \emph{knowledge-based} method of execution monitoring~\cite{pettersson2005execution} since we use data for learning, but also use an analytical approach for modelling the camera and body motion.
Several works use data-driven methods for execution monitoring and anomaly detection in robotics.
Park et al.~\cite{park2019multimodal, park2017multimodal} developed an execution monitor for detecting anomalous events during assistive tasks such as feeding.
They use force, sound and kinematic signals from a service robot to learn from nominal executions using hidden Markov models (HMMs) and Gaussian processes.
Inceoglu et al.~\cite{inceoglu2018failure} learn from several sensor modalities, using HMMs to classify extracted predicates from each modality into \emph{success} and \emph{failure} classes for different actions.
They also present an end-to-end convolutional neural network~\cite{inceoglu2020fino}, which classifies executions as \emph{success} or \emph{failure}, and identifies the failure types as well.
Wang et al.~\cite{wang2019multimodal} present a visual-tactile grasp dataset, consisting of tactile, joint and visual data of a robot arm grasping several objects.
Baseline results for slip detection using a recurrent neural network on the tactile data are presented.
\c{C}atal et al.~\cite{ccatal2020anomaly} present an anomaly detection method for an autonomous guided vehicle patrolling a warehouse, using a variational autoencoder to reconstruct an input image, conditioned on an action vector corresponding to commands sent to the robot.

All of these methods are characterised by the use of learning from multi-modal data using neural networks or hidden Markov models.
Colour and depth images are used in case of visual data, and kinematic data is represented as raw signals or processed to extract features.
In some cases, only nominal data is used for learning.
In our approach, we only focus on the \emph{motions} in the scene (represented by optical flow) since motion-related anomalies are more likely for tasks which involve interaction with the environment.
Instead of using raw kinematic signals, we represent them in a visual form, allowing us to model them in relation to the camera image.
Hence, our approach learns from nominal optical flow, and combines this with the kinematics of the robot, represented visually, without using learning.

\subsection{Video-based anomaly detection}
Reconstruction-based methods are one category of video anomaly detection approaches in which networks are trained to either reconstruct the input, or predict a future outcome, typically using some type of autoencoder network~\cite{ramachandra2020survey}.
These methods assume that the network will poorly reconstruct anomalous inputs since they are not in the same distribution as the training data.
In~\cite{liu2018future}, the authors use a U-Net network along with a discriminator for predicting a future frame.
They use intensity, gradient and optical flow loss between the predicted and ground truth frame, in addition to the discriminator loss to train the network.
In~\cite{gong2019memorizing}, the authors augment their autoencoder model with a memory module, which is updated with the latent codes of nominal videos during training.
At test time, inputs are reconstructed using the latent code in memory which is closest to the latent code of the test input.
Other variations of reconstruction-based approaches include using networks for both prediction and reconstruction~\cite{tang2020integrating}, reconstructing the input and predicting optical flow~\cite{nguyen2019anomaly}, the use of LSTMs in a variational autoencoder network~\cite{lu2019future} for future frame prediction, and the use of a graph-convolutional network to model object interactions~\cite{haresh20iv}.
We follow a similar approach to several of these methods by using a variational U-Net model to predict a future optical flow image.
However, in contrast to most methods, we only use a single optical flow image as input, instead of a sequence of RGB images, since we are primarily interested in modelling the motion as opposed to the appearance of the scene.
This makes the learning task less complex, and training and inference are faster due to lower input dimensionality.

\section{METHOD}
\begin{figure*}[thpb]
   \centering
   \includegraphics[width=0.99\textwidth]{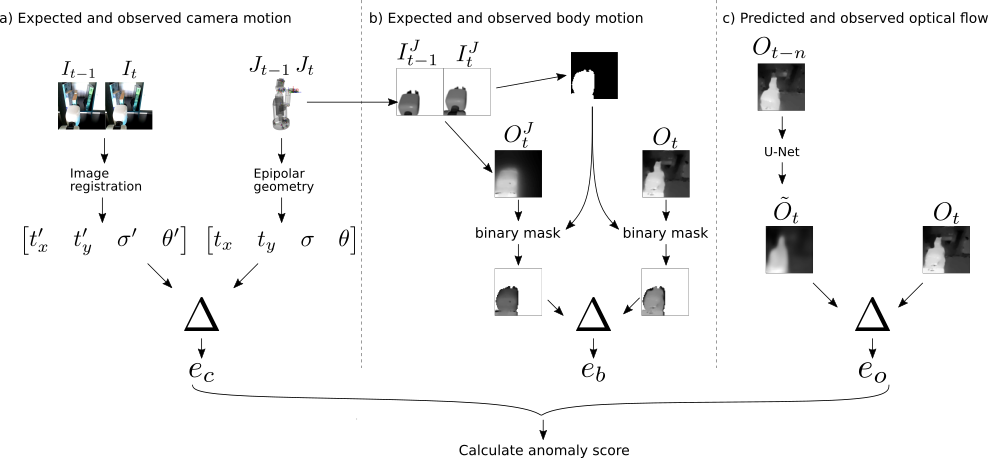}
   \caption{The expected motion is compared to the observed motion for three types of motions: a) \textbf{Camera motion:} the transformation between images obtained via image registration is compared with the expected transformation from known camera motion obtained from the robot's joint states b) \textbf{Body motion:} the observed optical flow of the robot's body (\(O_t\)) is compared with the optical flow obtained by rendering the robot model (\(O^J_t\)) using the robot's joint states c) \textbf{Optical flow:} the full optical flow image is compared to the optical flow image predicted by a trained U-Net network, given an optical flow image from the past. An anomaly score is calculated based on the residuals in each case (see text for notation).}
   \label{fig:overallarch}
\end{figure*}
Our goal is create a model of the nominal motion that occurs during the execution of a task, and to detect anomalies by monitoring the deviation from the nominal motion model.
By observing the \emph{motion}, we focus naturally on the salient regions of the scene - particularly those that are relevant for motion-related anomalies.
Motions of the robot's camera and body can be modelled using the internal sensors of the robot, and are less affected by the dynamics of the external environment in nominal scenarios.
We therefore use an optical flow prediction network to model the overall motion, but also model the camera and robot body motion separately using the robot's kinematics and joint states.
In all three cases, the error between the expected and observed motion is measured and used to calculate an anomaly score.

The sequence of images, \(I_{1:T}\), from the robot's camera and the robot state, \(J_{1:T}\), sampled at 10 Hz during the execution of the task are the inputs to the algorithms.
The robot state includes positions, velocities and efforts of all joints, and the frames of additional links of the robot.
Optical flow, which is calculated from consecutive pairs of images, is represented as a two-channel image (for horizontal and vertical displacement), \(O_{1:T}\).
Using the 3D model of the robot and the robot state, a sequence of images, \(I^J_{1:T}\), are rendered from the point of view of the robot's camera with the robot model in the configuration defined by the robot state.
An additional sequence of optical flow images, \(O^J_{1:T}\), are calculated from this sequence of rendered images.

Fig.~\ref{fig:overallarch} illustrates the three types of expected and observed motions which are compared.
For camera motion, we compare the \emph{expected} pixel motion calculated using the known camera motion against the \emph{measured} pixel motion using image registration.
For body motion, we compare the optical flow from \emph{rendered} images, \(O^J\), against the optical flow from \emph{real} images, \(O\).
For the optical flow, we use a probabilistic U-Net model~\cite{kohl2018probabilistic} to learn to predict the current optical flow, \(O_t\), given an optical flow frame from the past, \(O_{t-n}\).
The errors between the expected (or predicted) output and the measured output, \(e_c, e_b, \text{and } e_o\), are combined to calculate the anomaly score.
Each of the three errors are described in detail in the following sections.
\subsection{Camera Motion}
Motions of a robot-mounted camera cause an apparent motion of the entire scene.
The image motion caused by this camera motion is calculated in two ways.
\subsubsection{Expected Motion}
The camera motion relative to a fixed frame (such as the robot base) is obtained using the position sensors for various robot parts, such as the head and torso.
With epipolar geometry, the correspondence between image points in two consecutive views of the camera is obtained using Eq.~\ref{eq:generalmotion}~\cite[Eq. (9.7)]{hartley2003multiple}.
\begin{equation}
  \mathrm{x}' = K'RK^{-1}\mathbf{\mathrm{x}} + K'\mathbf{\mathrm{t}} / Z
  \label{eq:generalmotion}
\end{equation}
Here, \(\mathrm{x}'\) and \(\mathrm{x}\) are the image points in the two consecutive camera views corresponding to a point \(X\) in the scene, \(K = K'\) is the intrinsic camera matrix at the two time points, \(R\) and \(t\) are the rotation matrix and translation vector between the two camera positions and Z is the depth of the point \(X\) in 3-D space.
With a minimum of two such correspondences from the two views, the similarity transform (i.e. translation, \(t_x, t_y\), scale \(\sigma\), and rotation \(\theta\)) between the sets of image points is computed~\cite[pg. 39]{hartley2003multiple}.
In practice, more than two correspondences should be used to account for errors in the intrinsic parameters and noise in the depth data and proprioceptive sensors.
If the camera motion only consists of a translation, the simplified Eq.~\ref{eq:transmotion}~\cite[Eq. (9.6)]{hartley2003multiple} can be used instead.
\begin{equation}
  \mathrm{x}' = \mathbf{\mathrm{x}} + K'\mathbf{\mathrm{t}} / Z
  \label{eq:transmotion}
\end{equation}
In case no depth map of the scene is available, a uniform depth can be assumed for further simplification.
The RGB images from the camera are not used in this step.
\subsubsection{Observed Motion}
The observed motion between the images from the two viewpoints is computed using the Fourier-Mellin transform~\cite{reddy1996fft}.
This method registers two images by estimating the similarity transform, \([t'_x, t'_y, \sigma', \theta']\), between the images based on the Fourier shift theorem.
Since this method assumes that the dominant motion in the image is caused by camera motion, we mask out the region of the image where the robot body motion is visible before registration.
Finally, the absolute error between the expected and measured transforms is calculated.
In the use-case described in Sect.~\ref{sec:exp}, the camera motion results only in translations; hence we calculate the error as in Eq.~\ref{eq:cammotion}.
\begin{equation}
  e_c = |t_x - t'_x| + |t_y - t'_y|
  \label{eq:cammotion}
\end{equation}
For errors involving scale and rotation, a weighted sum of the errors should be used, or should be considered separately from the translation errors.
\subsection{Body Motion}
\label{sec:bodymotion}
\begin{figure}[thpb]
   \centering
   \includegraphics[width=0.99\linewidth]{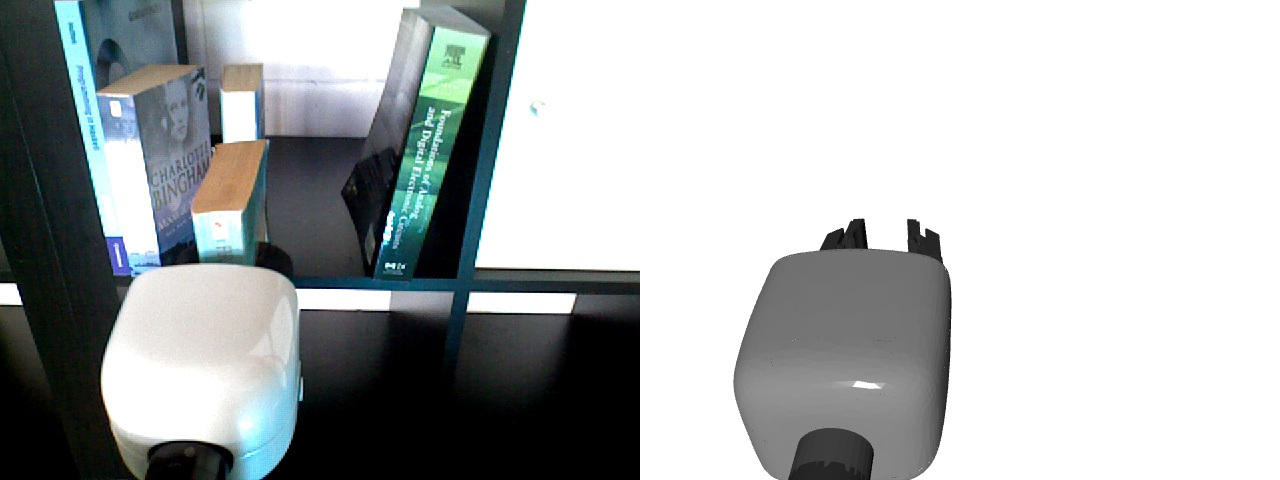}
   \caption{The real image (left) and rendered image (right) from the point of view of the robot's camera}
   \label{fig:pyrenderexample}
\end{figure}

The robot's manipulator and end-effector are typically in view of the camera during the execution of a manipulation task.
For the expected motion of the robot body, the robot model is rendered using the current robot state (\(J\)) and 3D model of the robot, and images (\(I^J\)) are captured from the pose of the real camera.
An example of the real and rendered image can be seen in Fig.~\ref{fig:pyrenderexample}.
The expected optical flow caused by the robot body, \(O^J\), is calculated from consecutive rendered frames, using the TV-L1 algorithm~\cite{zach2007duality}.

The optical flow between the real images, \(O\), is also calculated using the same algorithm.
The two optical flow images are masked such that only motions of the robot body remain, resulting in \(\bar{O}\) and \(\bar{O}^J\).
The mask is created by applying binary thresholding and contour detection on the rendered image of the robot body.
The error between the two optical flow images is calculated as the absolute difference between the median magnitude (denoted by \(med\))\footnote{We use the median instead of the mean to be robust to outliers} in both directions, as shown in Eq.~\ref{eq:bodymotion}.

\begin{equation}
  e_b = |med(\bar{O}^J_{x}) - med(\bar{O}_{x})| + |med(\bar{O}^J_{y}) - med(\bar{O}_{y})|
  \label{eq:bodymotion}
\end{equation}

\subsection{Optical Flow}
\begin{figure}[thpb]
   \centering
   \includegraphics[width=0.99\linewidth]{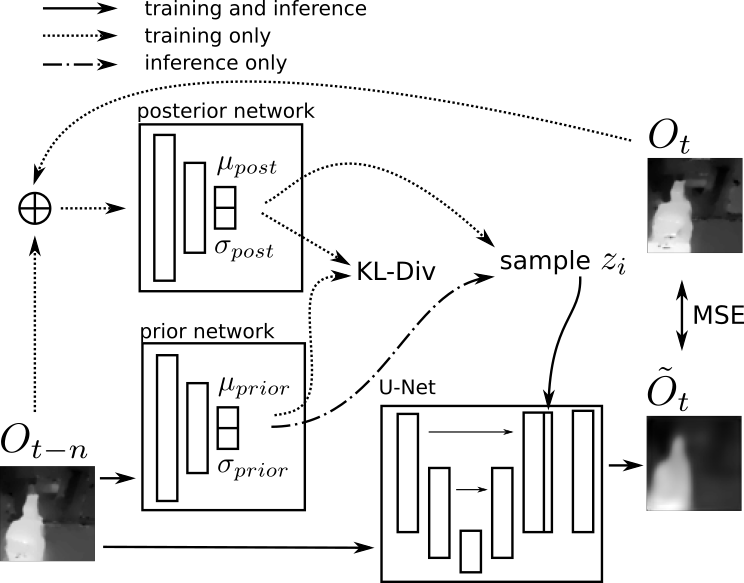}
    \caption{The probabilistic U-Net architecture consists of a posterior and prior network in addition to the standard U-Net. The network predicts a future optical flow image conditioned on a past optical flow image and a latent vector sampled from the distribution output by the posterior network (during training) or the prior network (during inference).}
   \label{fig:unet}
\end{figure}

To produce an expected optical flow image, we use a neural network which learns to predict the current optical flow image, \(O_t\), given a past optical flow image, \(O_{t-n}\).
We use a probabilistic U-Net~\cite{kohl2018probabilistic} model which combines a conditional variational auto-encoder (VAE) with a U-Net~\cite{ronneberger2015u}, by the addition of a prior and posterior network (see Fig.~\ref{fig:unet}).
This network architecture is chosen since it allows training with multiple ground truth outputs for a given input, and is relatively small compared to other frame prediction networks, with around 5 million trainable parameters.
In our case, we consider the prediction of an optical flow frame to be time-agnostic~\cite{jayaraman2019time}; the past optical flow frame which can best predict the current frame is selected from a range of past frames rather than from a fixed offset in the past.
During training, an optical flow image from the past, \(O_{t-n}\), is selected randomly within a certain range,  where \(n \in [A,...,A+B]\), with the target optical flow image as \(O_t\).
At inference time, however, \(\tilde{O}_t\) is predicted \(B\) times using inputs \(O_{t-n}\) for all \(n\), and the prediction with the lowest error is selected.
The lower limit of the range, \(A\), and the span of the range, \(B\), are hyperparameters whose values we determine experimentally.

The prior network produces a distribution \(P(z|O_{t-n})\) represented by the parameters of a multi-dimensional Gaussian with a diagonal covariance matrix (\(\mu_{prior}, \sigma_{prior}\)).
At inference time, a latent vector, \(z_i\), is sampled from this distribution, concatenated with the last activation map of the U-Net, and passed through some final convolutional layers to produce \(P(O_t | UNet(O_{t-n}), z_i)\) (where the input to the U-Net is also \(O_{t-n}\)).
Multiple \(\tilde{O}_t\) can be generated by sampling multiple times from the prior distribution.

During training, the latent vector is sampled from the distribution, (\(\mu_{post}, \sigma_{post}\)), parameterized by the posterior network, \(Q(z|O_t, O_{t-n})\), instead of the prior network.
The input to the posterior network is a concatenation along the channel dimension of the ground truth, \(O_t\), and the input to the prior network, \(O_{t-n}\).
The posterior network learns to produce a latent vector that will result in the exact ground truth provided in its input.
The training objective (Eq.~\ref{eq:loss}) is that of a VAE~\cite{kingma2013auto}, namely, the sum of mean-squared error (MSE) between the predicted output and the ground truth, and the Kullback-Leibler (KL) divergence between the distributions output by the posterior and prior networks.\footnote{The loss function is identical to that used in~\cite{kohl2018probabilistic} with the exception that the output distribution \(P(O_t)\) is considered to be a normal distribution instead of a categorical distribution}
The KL-divergence term, which is weighted by \(\beta\), a hyperparameter, encourages the prior distribution to move close to the posterior distribution.\footnote{We use \(\beta = 10.0\), which is the default in the implementation by~\cite{kohl2018probabilistic}}

\begin{multline}
  \mathcal{L}(O_t, O_{t-n}) = \\\mathbb{E}_{z\sim Q(\cdot|O_t,O_{t-n})}[-\text{log}P(O_t|UNet(O_{t-n}),z)]\\+ \beta \cdot D_{KL}(Q(z|O_t,O_{t-n})||P(z|O_{t-n}))
  \label{eq:loss}
\end{multline}

The input optical flow images are resized and center-cropped to 64x64 pixels.
All models are trained for 50 epochs using a batch size of 128 and learning rate of 0.0001.
At inference time, we sample \(M\) outputs from the network for each \(n\) (resulting in a total of \(M \times B\) predictions), and the output, \(\tilde{O}_t\), with the minimum prediction error is used to compute the error as in Eq.~\ref{eq:eo}.
\begin{equation}
  \label{eq:eo}
  e_o = min_{n \in [A,...,A+B]} min_{j \in [1..M]} (mse(O_t, \tilde{O}^{nj}_{{t}}))
\end{equation}
\subsection{Anomaly Score}
We do not expect a significant change in \(e_c\) and \(e_b\) between the training and nominal test data since they are based on the robot's proprioceptive sensors.
However, we do expect a change in \(e_o\) at inference time due to distribution shift, even for nominal executions.
Therefore, we use a fixed threshold for the camera and body motion, but evaluate the overall score using a range of thresholds.
The thresholds for the camera and body motion errors are based on the maximum error from the training set, as in Eq.~\ref{eq:cbthresh}.
\begin{equation}
  \label{eq:cbthresh}
  \begin{aligned}
    e_{c_{thres}} = \text{max}_{train}(e_c) \\
    e_{b_{thres}} = \text{max}_{train}(e_b)
  \end{aligned}
\end{equation}

The final anomaly score is calculated as:
\begin{equation}
\label{eq:score}
  \text{score} =
\begin{cases}
  1,& \text{if } e_c\geq e_{c_{thres}} \text{ or } e_b\geq e_{b_{thres}}\\
  e_o  ,              & \text{otherwise}
\end{cases}
\end{equation}
%

\section{EXPERIMENTS}
\label{sec:exp}
\subsection{Data}
The method is evaluated on a dataset of executions in which the robot places a book on a shelf.
The dataset consists of 61 nominal executions and 60 anomalous executions.
Data recorded includes RGB and depth images from the head-mounted 3-D camera, joint states, and other sensor data such as the force-torque sensor.
Anomalies include the book falling on or off the shelf, books on the shelf being disturbed significantly, occluded camera, and external collisions and disturbances to the robot.
While camera occlusions and disturbances to the robot do not necessarily result in task failures, they are still important to detect since they might indicate other problems that the robot needs to address (for example, the robot may need to re-localize itself if it has been disturbed).
Anomalies are annotated frame-wise, so that the objective is to detect anomalous frames.
For training the learning model and for determining thresholds for the camera and body motion models, 48 nominal executions are used as training data and an additional 6 nominal executions are used for validation.
The test set consists of 60 anomalous executions and 7 nominal executions.

\subsection{Evaluation Metrics}
\label{sec:metrics}
The area under the receiver operating characteristic curve (AUC-ROC) is the most common metric used in video anomaly detection.
It allows us to compare methods based on their ability to discriminate between anomalous and non-anomalous frames at different thresholds.
If there is an imbalance between the two classes, the AUC of the precision-recall (AUC-PR) curve is an additional metric which summarizes how well the classifier is able to detect the anomalous frames at different thresholds.
Since the dataset is unbalanced (only about 12\% of test frames are anomalous), both metrics are used for evaluation.
In both cases, an area of 1.0 represents a perfect detector.
Unlike several video-anomaly detection methods (see~\cite{ramachandra2020survey}), we do not normalize the anomaly score per execution, since this assumes that at least one anomaly occurs during an execution.

\subsection{Results}

\begin{figure*}[thpb]
   \centering
   \includegraphics[width=0.97\textwidth]{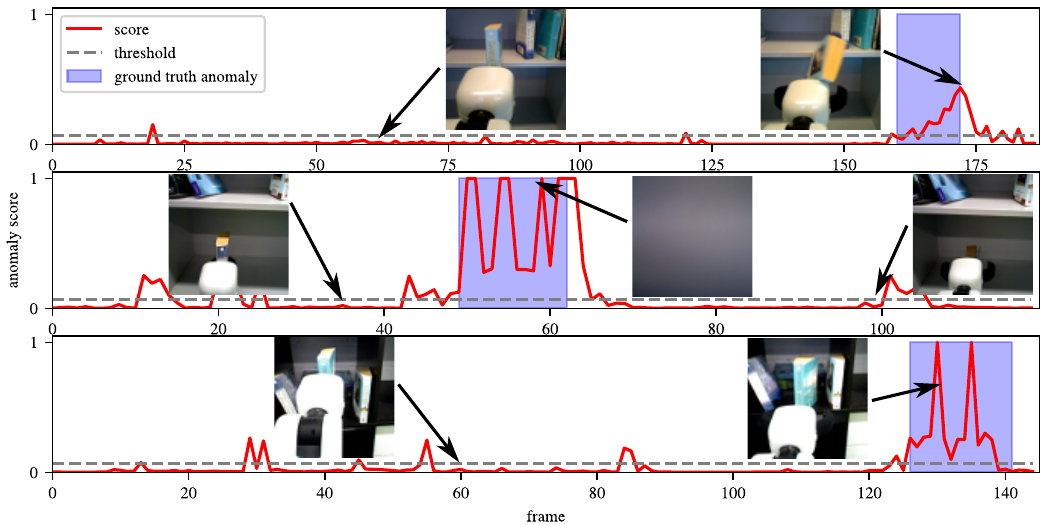}
   \caption{Several types of anomalies are illustrated here: \textbf{top:} the book starts to falls off the shelf around frame 160 when the arm is retracted; \textbf{middle:} the camera is occluded at around frame 50; \textbf{bottom:} the robot is disturbed externally, resulting in a shaking camera around frame 130. The threshold shown here corresponds to the optimal point in the precision-recall curve. The full clips for these examples can be found in the supplementary video.}
   \label{fig:results}
\end{figure*}
\subsubsection{Comparison to other anomaly detection methods}
We compare the performance of the probabilistic U-Net to two other future prediction methods, U-Net + GAN~\cite{liu2018future} and VRNN~\cite{lu2019future}, and an HMM-based method similar to~\cite{park2019multimodal}.
We train the probabilistic U-Net with \(n = 1\), such that it only predicts the next optical flow frame.
For a fair comparison with the future prediction models, we do not consider the body and camera motion, so that the anomaly score is \(e_o\).
Both future prediction models are trained to predict the next RGB frame given a sequence of input RGB frames.
We fit the HMM using the Baum-Welch algorithm with multivariate features comprising of the maximum magnitude from optical flow images and magnitudes of observed body motion and image motion due to camera motion.
The anomaly score for each frame in the test set is computed using the log likelihood of the sequence of observations up to that frame.
The results in Table~\ref{tbl:comparison} show that learning to predict optical flow using the probabilistic U-Net has an advantage over the future prediction methods, and the HMM.
\begin{table}[h]
  \caption{Comparison to other methods}
  \label{tbl:comparison}
  \centering
    \begin{tabular}{ccc}
    \hline
      Method & AUC-ROC & AUC-PR \\
    \hline
      U-Net + GAN~\cite{liu2018future} & 0.675 & 0.339 \\
      VRNN~\cite{lu2019future} & 0.533 & 0.166\\
      HMM & 0.657 & 0.197 \\
      Prob. U-Net & \textbf{0.728} & \textbf{0.397}\\
    \hline
  \end{tabular}
\end{table}
\subsubsection{Input range}
\begin{figure}[thpb]
   \centering
   \includegraphics[width=0.85\linewidth]{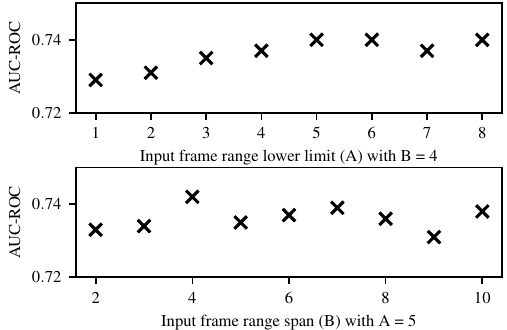}
   \caption{AUC-ROC for different ranges of input optical flow images}
   \label{fig:Aplot}
\end{figure}

We determine ideal values for the upper and lower limit of the range (\(A, A+B\)) of past frames to be used as the input to the probabilistic U-Net.
As seen in Fig.~\ref{fig:Aplot}, there is a marginal improvement of AUC-ROC as the lower limit and span are increased up to a certain point.
Similar trends were observed for other combinations of \(A\) and \(B\), and for AUC-PR.
We obtain the best results for \(A=5\) and \(B=4\), namely by using an optical flow frame between 5 and 9 time steps in the past to predict the current frame.

\subsubsection{Optical flow, body and camera motion}
For the learning method, we consider several variants of the input optical flow.
The \emph{registered optical flow} is the optical flow calculated after registering consecutive images.
In effect, the camera motion is no longer visible in this variant.
The \emph{masked optical flow} masks out the optical flow in the regions where the robot's body is visible (using the inversion of the mask described in~\ref{sec:bodymotion}).
This removes observed motions of the robot body from the optical flow.
The \emph{masked registered} variant is a masked version of the registered optical flow.
Table~\ref{tbl:resultsroc} shows the AUC-ROC and AUC-PR for all optical flow variants using 
\begin{inparaenum}[\itshape (i)\upshape]
  \item only the learning method (OF only; \(\text{score} = e_o\));
  \item the learning method + error from the body motion (OF + body motion); and
  \item the learning method combined with the body and camera motion errors (OF + body + camera motion; score calculated as in Eq.~\ref{eq:score}).
\end{inparaenum}
All results use \(A=5\), \(B=4\) and \(M=10\).
\begin{table}[thpb]
  \caption{AUC-ROC and AUC-PR}
  \label{tbl:resultsroc}
  \centering
    \begin{tabular}{p{0.16\textwidth}p{0.07\textwidth}p{0.07\textwidth}p{0.07\textwidth}}
    \hline
      Input optical flow variant & OF only & OF + body motion & OF + body + camera motion\\
    \hline
                                                     &  \multicolumn{3}{c}{AUC-ROC}\\
    \cline{2-4}
      Optical flow                                   &  0.740 & 0.741 & 0.792\\
      Registered optical flow                        &  0.716 & 0.716  & 0.773\\
      Masked optical flow                            &  \textbf{0.752} & \textbf{0.752} & \textbf{0.804}\\
      Masked reg. optical flow                       &  0.727 & 0.727  & 0.779\\
    \hline
                                                     &  \multicolumn{3}{c}{AUC-PR}\\
    \cline{2-4}
      Optical flow                                   &  0.380 & 0.387 & 0.489  \\
      Registered optical flow                        &  0.353 & 0.362  & 0.466 \\
      Masked optical flow                            &  \textbf{0.467} & \textbf{0.464} & \textbf{0.549}  \\
      Masked reg. optical flow                       &  0.413 & 0.415 & 0.509 \\
    \hline
  \end{tabular}
\end{table}

The masked optical flow variant performs the best in all cases.
Incorporating the body motion error has no effect on the AUC-ROC, and only marginally improves AUC-PR.
However, incorporating the camera motion error shows an improvement in both metrics.
Fig.~\ref{fig:results} shows the anomaly score with some corresponding image frames.
Unexpected motions such as a falling book (top), and shaking of the camera (bottom) result in an increase in the anomaly score.
Occlusion of the camera (middle) also results in a high anomaly score.
\subsection{Discussion}
The body motion error did not impact the performance significantly; this is probably because the types of anomalies in the dataset did not include discrepancies between the internally sensed arm motion and the observed motion.
Body motion errors were visible when, for example, the book occluded the view of the arm, or the camera was occluded.

The camera motion error significantly improved the performance in terms of detecting the anomalies when they occurred (AUC-PR).
In addition to detecting disturbances to the robot and occlusions, anomalies which involved large motions also increased the detection rate since they affected the image registration process (which assumes that the dominant motion in the scene is due to camera motion).
Using registered optical flow decreased performance in all cases; this is likely due to the same reason.

The motions of the manipulator probably made the task of predicting the motion harder, since they caused significant motion in the image due to being close to the camera.
Therefore, using the masked optical flow performed better.
The method failed in cases of static anomalies (such as a book \emph{nearly} falling out of the gripper), and instances where the anomaly was mostly occluded by the arm.
False positives were seen in cases of inaccurate optical flow calculation, and often at the time of release of the book.
Our dataset does not include background motions (such as persons moving around) which are unrelated to the task.
This is an aspect that should be evaluated in future work, since unstructured motions in the background are likely to make predicting future motions harder.

\section{CONCLUSIONS}
We investigated using visual anomaly detection for monitoring the execution of tasks by modelling the motions observed during successful executions.
We use a combination of a learning and model-based method to generate expectations about the nominal motion, which are compared to the observed motions to detect anomalies.
Our experiments show that it is beneficial to separately consider the known motions of the robot (in particular, the camera motion) when comparing the expected and observed motions.
The results also improve if the motions of the robot body are removed from the input to the learning method.
Our approach incorporates the robot's kinematics and model as visual inputs to the anomaly detection method.
However, non-visual task knowledge, such as task progress, could provide additional context and structure to the neural network or to the overall method, and is a good candidate for future research.


\end{document}